\documentclass[letterpaper, 10 pt, conference]{ieeeconf}  

\IEEEoverridecommandlockouts                              

\usepackage{graphics} 
\usepackage{times} 
\usepackage{amsmath} 
\usepackage{amssymb}  
\usepackage{graphicx}
\usepackage{color}
\usepackage{cite}
\makeatletter
\let\NAT@parse\undefined
\makeatother
\usepackage[colorlinks,linkcolor=blue, citecolor=blue, anchorcolor=blue]{hyperref}
\usepackage{multirow}

\title{\LARGE \bf
Caterpillar-Inspired Spring-Based Compressive Continuum Robot for Bristle-based Exploration}

\author{Zhixian Hu, Yu She, and Juan Wachs
\thanks{This material is partially based upon work supported by the National Science Foundation under Grant NSF \#2140612, 
NSF \#2423068, and Showalter Trust. Any opinions, findings, and conclusions or recommendations expressed in this material are those of the author(s) and do not necessarily reflect the views of the funding agencies.}
\thanks{Edwardson School of Industrial Engineering, Purdue University, West Lafayette, IN, USA. \tt\small{\{shey, jpwachs\}@purdue.edu}}
}

\begin{document}

\maketitle
\thispagestyle{empty}
\pagestyle{empty}

\begin{abstract}
Exploration of confined spaces, such as pipelines and ducts, remains challenging for conventional rigid robots due to limited space, irregular geometry, and restricted access. Inspired by caterpillar locomotion and sensing, this paper presents a compact spring-based tendon-driven continuum robot that integrates with commercial robotic arms for confined-space inspection. The system combines a mechanically compliant continuum body with a tendon actuation module, enabling coupled bending and axial length change, and uses a constant-curvature kinematic model for positional control. Experiments show a mean position error of 4.32 mm under the proposed model and control pipeline. To extend the system from motion to inspection, we integrate an artificial bristle contact sensor and demonstrate surface perception and confined-space exploration through contact interactions. This compact and compliant design offers a cost-effective upgrade for commercial robots and promises effective exploration in challenging environments.

\end{abstract}

\section{INTRODUCTION}

Deformable bodies in animals enable robust interaction with dynamic environments. Soft robotics draws on this idea to provide versatile and resilient mechanisms for exploration and manipulation. Along this line, continuum robots have been developed by borrowing concepts from caterpillars, elephant trunks, and octopus tentacles \cite{martinez2013robotic, xie2020proprioceptive, taylor2023tactile, paek2015microrobotic, lee2016elastic, lin2023recent}. A central goal is to achieve high motion flexibility for navigation, inspection, and manipulation in confined spaces, including pipelines, ventilation ducts, reactor vessels, and human body cavities. Such systems support minimally invasive surgery \cite{lee2016elastic}, environment characterization \cite{zou2018reconfigurable, umedachi2019caterpillar}, and machine inspection \cite{wang2021bio}. 

\begin{figure}[thpb]
  \centering
  \includegraphics[width=0.485\textwidth]{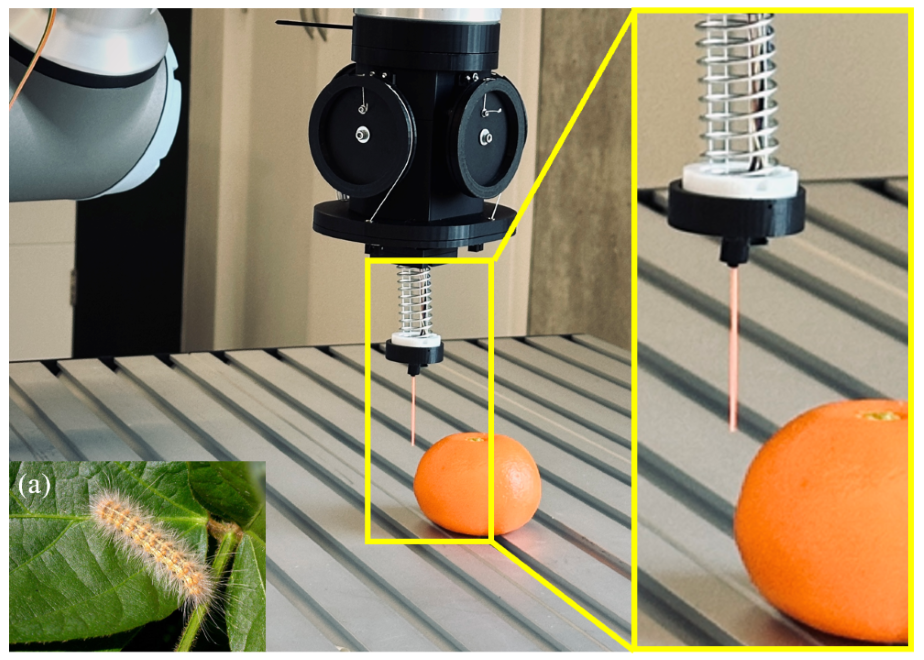}
  \caption{Overview of the bristled-caterpillar-inspired robot. (a) Reproduced under the terms of the \href{https://creativecommons.org/licenses/by-nc/2.0/}{CC BY-NC 2.0 DEED}, Copyright 2006, \href{https://flickr.com/photos/57402879@N00/199014708/in/photolist-iA195-XBMugW-Hk9EoV-2YgME-9UgrQU-pzrAsT-vjUXH5-4XKR6-9Xqmgd-6LsEWi-6Nqz72-y9FtiR-5RAjAB-4MupQV-zsUiB2-AX73Eb-k6FnB-9ePRL6-9HbEgr-6x3dcY-7HFKXb-2zuVR-4LpRGV-3Bh7b-8qAfTg-5jtSj9-4ZXh8-jx1q6-RPLJx-6SMXqw-deXymu-7Lqjsf-XXX9o3-27SW8ZG-opve95-g9kAVm-7bpX6L-8QPRwW-3EEk3-4Dqscm-aSb2f-BaQYm-9oB93z-4EZFaY-Ybg5TR-7oxXU8-9de9g6-aJSmpz-a35Kh3-7HUBdq}{Brad Smith / flickr}. }
  \label{intro}
\end{figure}

A major research focus is reproducing compressive motions in continuum robots, which are characteristic of caterpillars and other animals. Common actuation methods are pneumatic-based systems, concentric tube transmission, magnetic-driven mechanisms, and tendon-driven approaches\cite{zhong2020recent}. Pneumatic-based robots utilize either everted tube\cite{hawkes2017soft, coad2020retraction} or pressurized cavity\cite{chen2019design, zou2018reconfigurable, sheng2020multi}, offering low mass and fast response\cite{zhong2020recent}. However, besides the necessity of bulky pumps as actuating devices, the former design typically have low payload, and the latter have limited workspace due to cavity shapes. Concentric tube robots use nested pre-curved tubes and enable compact miniaturized designs\cite{nwafor2023design, gunderman2022surgical}, but they can suffer from reduced flexibility and increased complexity in kinematics and control. Magnetic-driven continuum robots embed magnetic components in compressive robot bodies (e.g., origami structures) and are driven by the change of magnetic field\cite{cai2020diversified}. However, the need for external field generation limits practical deployment.

Tendon-driven continuum robots are a promising alternative. They typically use a central backbone segmented for bending, where pulling tendons produces bending or twisting. This design supports interactions with constrained environments\cite{zhong2020recent} and enables relatively easy kinematics and control. Accompanying compressive section designs, tendon-driven robots become suitable for in-space inspection and exploration. Umedachi et al. designed a robot with deformable beams that compress and bend concurrently, but uniaxial bending limits flexibility\cite{umedachi2019caterpillar}. Nguyen and Burgner-Kahrs proposed a telescoping backbone with magnetic spacer disks\cite{nguyen2015tendon}. Magnetic repulsion distributes the disks along the backbone, enabling section-length adjustment and shape adaptation through tendon force control. Xu et al. utilized a bidirectional asymmetric V-shaped notched continuum structure for compressive motion\cite{xu2024novel}, and demonstrated its use in a phantom experiment for vocal fold tissue examination.

Compression springs have also been explored as compliant backbones for tendon-driven robots with compressive features. Li et al. designed a two-module continuum robot with helical springs\cite{li2018design}, focusing primarily on bending control rather than coupled extending and bending. A miniature spring-based manipulator was proposed for bronchoscopic steering by Wang et al. \cite{wang2023preliminary}, while the design of only two parallel tendons restricts workspace. In addition, many spring-based designs use actuation modules that are large relative to the manipulator body, which limits integration and deployment\cite{kim2017active, wang2025design}.

This paper proposes a spring-based tendon-driven continuum robot inspired by caterpillar compressive bending, targeting flexible exploration in confined environments (Fig. \ref{intro}). The compact actuation module enables integration with commercial robotic arms as an end-effector. Additionally, combined with an artificial bristle, the manipulator can perform in-space inspection tasks utilizing its compliant and non-invasive structure. 

\section{METHOD}
The caterpillar-mimicked continuum robot comprises a manipulation module and a perception module. The manipulation module uses tendon actuation with a compression spring backbone, enabling both bending and compressing motions. The perception module is an artificial bristle mounted at the tip to detect contacts. This combination of movement and sensing allows the robot to explore the environment like a caterpillar.

\subsection{Mechanical Design of the Manipulation Module}

\begin{figure}[thpb]
  \centering
  \includegraphics[scale=0.65]{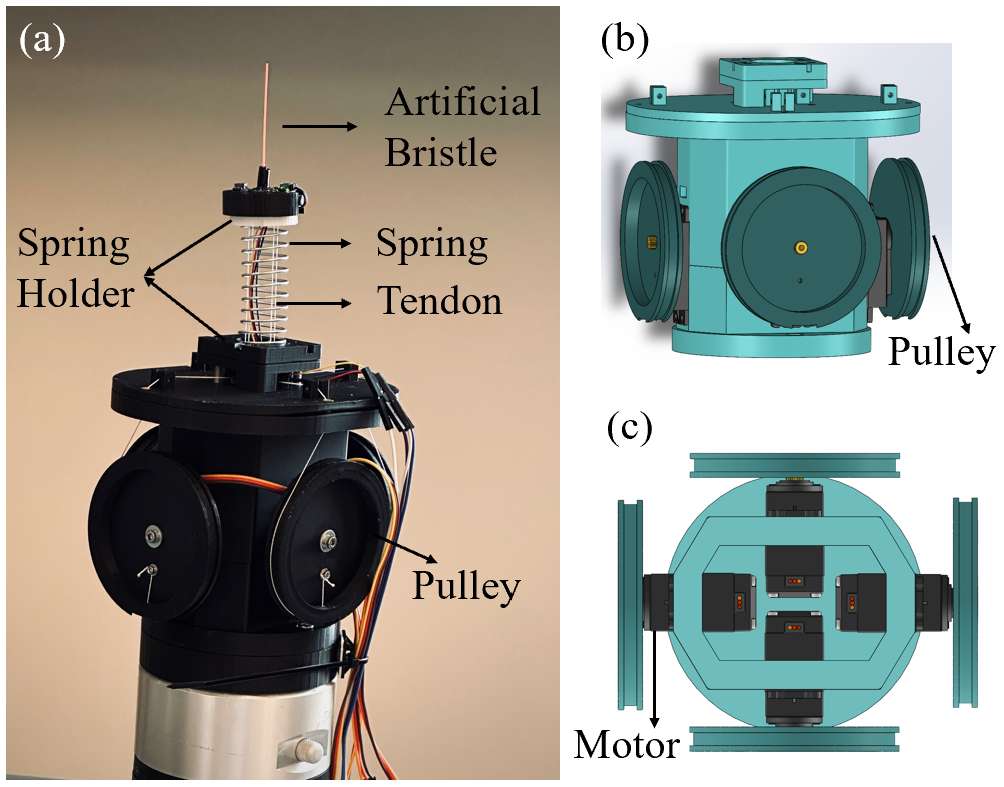}
  \caption{Mechanical design. (a) An overview of the mechanical design of the continuum robot. (b) and (c) are the side view and the top sectional view of the actuation part of the continuum robot. }
  \label{struct}
\end{figure}

The manipulation module comprises an actuation part and a manipulator part (Fig. \ref{struct}). The actuation part is a four-tendon system. Each tendon is anchored to the top spring holder and routed to a servo motor via a circular pulley. Tendon length is controlled by rolling the pulley on the servo, which drives the manipulator motion. Fig. \ref{struct} (b) and (c) show a holding base housing the motors and the manipulator, ensuring the robot's compactness and compatibility with commercial robotic systems. The holding-base components are 3D-printed. The manipulator uses a compressing spring as the backbone. Coordinated actuation of the four tendons enables bending in arbitrary directions as well as axial compression. The spring elasticity also permits bending at different effective lengths, producing a range of bending angles and orientations.

\subsection{Kinematics Modeling of the Manipulation Module}

\begin{figure}[thpb]
  \centering
  \includegraphics[scale=0.43]{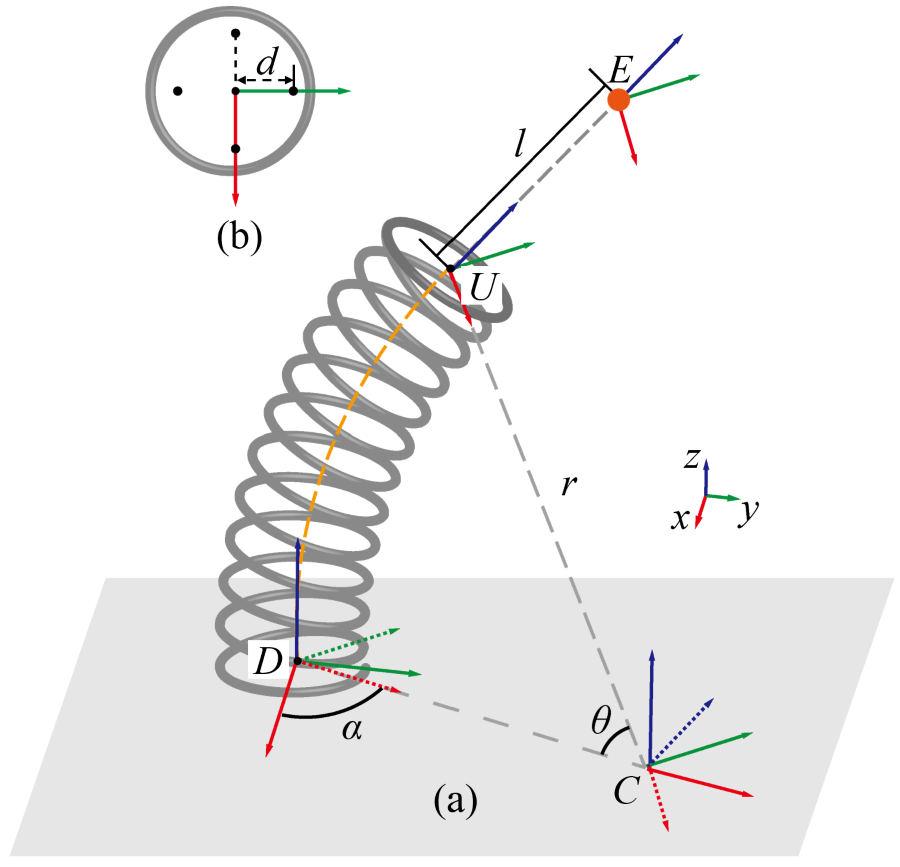}
  \caption{(a) Kinematic modeling of the continuum robot. The x axes of the coordinate frames are denoted by the red arrows, the y axes are green, and the z axes are blue. (b) The relationship between the tendon-attached points and the spring holders.}
  \label{knm}
\end{figure}

A constant curvature kinematic model is used as a mapping from tendon length commands to robot pose, as commonly adopted for continuum robots\cite{webster2010design, camarillo2008mechanics}. For a compliant, axially compressible spring backbone, this assumption is approximate and does not explicitly model distributed deformation or tendon effects (e.g., friction and backlash), which cause region-dependent deviations as quantified in the error analysis. Several assumptions are made for this kinematic analysis. The spring centerline can be conceptualized as a circular arc with arc length that varies with tendon adjustment. The bending body is modeled as a torus segment in a geometric framework, from which the kinematic model for the spring-based robot is constructed (Fig. \ref{knm}).

Denote the center of the spring bottom as $D$, the center of the spring top as $U$, and the center of the corresponding circular arc as $C$. The coordinate systems are defined in Fig. \ref{knm} (a). The homogeneous transformation matrix from Frame $D$ to Frame $C$ is\begin{equation} \label{DTC} 
\textsuperscript{D}T_C = Rot_z(\alpha)D_x(r),\end{equation}
where $Rot_{s}(g)$ is the rotation about the $s$ axis by $g$ degree, $D_{s}(k)$ is the translation along the $s$ axis by distance $k$, $\alpha$ is the bending angle of the spring along the z-axis of Frame $D$, and $r$ is the radius of the circular arc. 

The homogeneous transformation matrix from Frame $C$ to Frame $U$ is
\begin{equation}\label{CTU}
\textsuperscript{C}T_U = Rot_y(\theta)D_x(-r)Rot_z(-\alpha),
\end{equation}
where $\theta$ is the bending angle about arc center.

Combining (\ref{DTC}) and (\ref{CTU}), the homogeneous transformation matrix from Frame $D$ to Frame $U$ is\\
$\textsuperscript{D}T_U =\textsuperscript{D}T_C\textsuperscript{C}T_U$
\begin{equation}\label{DTU}
=\begin{bmatrix}
    c_{\alpha}^2c_{\theta}+s_{\alpha}^2 & s_{\alpha}c_{\alpha}c_{\theta}-s_{\alpha}c_{\alpha} & c_{\alpha}s_{\theta} & -rc_{\alpha}c_{\theta}+rc_{\alpha} \\
    s_{\alpha}c_{\alpha}c_{\theta}-s_{\alpha}c_{\alpha} & s_{\alpha}^2c_{\theta}+c_{\alpha}^2 & s_{\alpha}s_{\theta} & -rs_{\alpha}c_{\theta}+rs_{\alpha} \\
    -c_{\alpha}s_{\theta} & -s_{\alpha}s_{\theta} & c_{\theta} & rs_{\theta}\\
    0 & 0 & 0 & 1
\end{bmatrix},
\end{equation}
where $c_{\alpha}$ and $c_\theta$ are short for $cos(\alpha)$ and $cos(\theta)$, and $s_\alpha$ and $s_\theta$ are short for $sin(\alpha)$ and $sin(\theta)$.

Given the position of point $U$ to be $(x_U, y_U, z_U)$,
\begin{equation}
    \left\{\begin{aligned}
    & x_U =  -rc_{\alpha}c_{\theta}+rc_{\alpha},\\ 
    & y_U = -rs_{\alpha}c_{\theta}+rs_{\alpha},\\
    & z_U =  rs_{\theta}.\\
    \end{aligned}
    \right. \label{xyzU}
\end{equation}

By solving (\ref{xyzU}), 
\begin{equation}\label{theta}
\theta = arccos(\frac{z_U^2-x^2_U-y^2_U}{x^2_U+y^2_U+z^2_U}),
\end{equation}
\begin{equation}\label{alpha}
\alpha = arctan(\frac{y}{x}),
\end{equation}
\begin{equation}\label{r}
r=\frac{x_U^2+y_U^2+z_U^2}{2\sqrt{x_U^2+y_U^2}},
\end{equation}
with $\theta\in[0,\frac{\pi}{2}]$, $sin(\alpha) = \frac{y}{\sqrt{x^2+y^2}}$ and $cos(\alpha)=\frac{x}{\sqrt{x^2+y^2}}$.

For the tip $E$ of the robot, the homogeneous transformation matrix from Frame $U$ to Frame $E$ is
\begin{equation}\label{UTE}
\textsuperscript{U}T_E = D_z(l),
\end{equation}
where $l$ is the distance between the center of the top of the spring and the robot tip. Then the homogeneous transformation matrix from Frame $D$ to Frame $E$ is

$\textsuperscript{D}T_E =\textsuperscript{D}T_U\textsuperscript{U}T_E$\\
\begin{equation}\label{DTE}
=\begin{bmatrix}
    c_{\alpha}^2c_{\theta}+s_{\alpha}^2 & s_{\alpha}c_{\alpha}c_{\theta}-s_{\alpha}c_{\alpha} & c_{\alpha}s_{\theta} & lc_{\alpha}s_\theta+x_U \\
    s_{\alpha}c_{\alpha}c_{\theta}-s_{\alpha}c_{\alpha} & s_{\alpha}^2c_{\theta}+c_{\alpha}^2 & s_{\alpha}s_{\theta}& ls_\alpha s_\theta+y_U \\
    -c_{\alpha}s_{\theta} & -s_{\alpha}s_{\theta} & c_{\theta} & lc_\theta +z_U\\
    0 & 0 & 0 & 1
\end{bmatrix},
\end{equation}

Thus, the position of the robot tip $E$ could be obtained by
\begin{equation}
    \left\{
    \begin{aligned}
    & x_E =  lc_{\alpha}s_{\theta}+x_U,\\
    & y_E = ls_{\alpha}s_{\theta}+y_U,\\
    & z_E =  lc_{\theta}+z_U.\\
    \end{aligned}
    \right. \label{xyzE}
\end{equation}

Because the continuum robot is tendon-driven, the length of each tendon needs to be computed. From Fig. \ref{knm} (b), we could obtain the positions of the tendon-attached points of the lower part are 
\begin{equation}
    \left\{
    \begin{aligned}
    & P_{L,1} =  (d, 0, 0),
    & P_{L,2} = (0, d, 0),\\
    & P_{L,3} =  (-d, 0, 0),
    & P_{L,4} = (0, -d, 0).
    \end{aligned}
    \right.
    \label{lowP}
\end{equation}

Combined with (\ref{DTU}), the positions of the tendon-attached points of the higher part are 
\begin{equation}
    \left\{
    \begin{aligned}
    & P_{H,1} =  (dc_{\alpha}^2c_{\theta}+ds_{\alpha}^2+x_U, ds_{\alpha}c_{\alpha}c_{\theta}-ds_{\alpha}c_{\alpha}+y_U, \\
    &\quad\quad-dc_{\alpha}s_{\theta}+z_U),\\
    & P_{H,2} = (ds_{\alpha}c_{\alpha}c_{\theta}-ds_{\alpha}c_{\alpha}+x_U, ds_{\alpha}^2c_{\theta}+dc_{\alpha}^2+y_U, \\
    &-ds_{\alpha}s_{\theta}+z_U)\\
    & P_{H,3} =  (-dc_{\alpha}^2c_{\theta}-ds_{\alpha}^2+x_U, -ds_{\alpha}c_{\alpha}c_{\theta}+ds_{\alpha}c_{\alpha}+y_U, 
    \\&dc_{\alpha}s_{\theta}+z_U),\\
    & P_{H,4} =(-ds_{\alpha}c_{\alpha}c_{\theta}+ds_{\alpha}c_{\alpha}+x_U, -ds_{\alpha}^2c_{\theta}-dc_{\alpha}^2+y_U, \\&ds_{\alpha}s_{\theta}+z_U).
    \end{aligned}
    \right. \label{highP}
\end{equation}

Tendon length computation is divided into two cases, depending on whether the tendon attachment point lies on the inner semicircle of the spring (closer to $C$) or the outer semicircle (farther from $C$). In the inner case, the tendon is approximated as a circular arc with radius $||P_{L,i}-C||_2$ ($i=1,2,3,4$). In the outer case, the tendon is modeled as a straight link. Given the desired position of the spring top center $U$, the four tendon lengths $q_i$ ($i=1,2,3,4$) are computed as follows:
\begin{equation}
    q_i=\left\{
    \begin{aligned}
    & ||P_{L,i}-C||_2\cdot\theta,\text{ when }||P_{L,i}-C||_2<\sqrt{r^2+d^2},\\
    & ||P_{L,i}-P_{H,i}||_2,\text{ when }||P_{L,i}-C||_2\geq \sqrt{r^2+d^2},
    \end{aligned}
    \right. \label{tendonLength}
\end{equation}
where $||\cdot||_2$ denotes the Euclidean norm.


\subsection{Design of the Bristle Perception Module}

In this paper, we adopt the artificial bristle design by Xiao and Wachs \cite{xiao2023complacent, Xiao2023} for contact detection. As shown in Fig. \ref{struct} (a), a plastic filament bristle is bonded to a pressure sensor through silicone. Contact with the environment induces a pressure change that is used to detect contact events. The bonding silicone is Ecoflex 00-30 (Smooth-On, Inc.). 

\section{EXPERIMENTS}
\subsection{Experiment Setup}

\begin{figure}[t]
  \centering
  \includegraphics[scale=0.7]{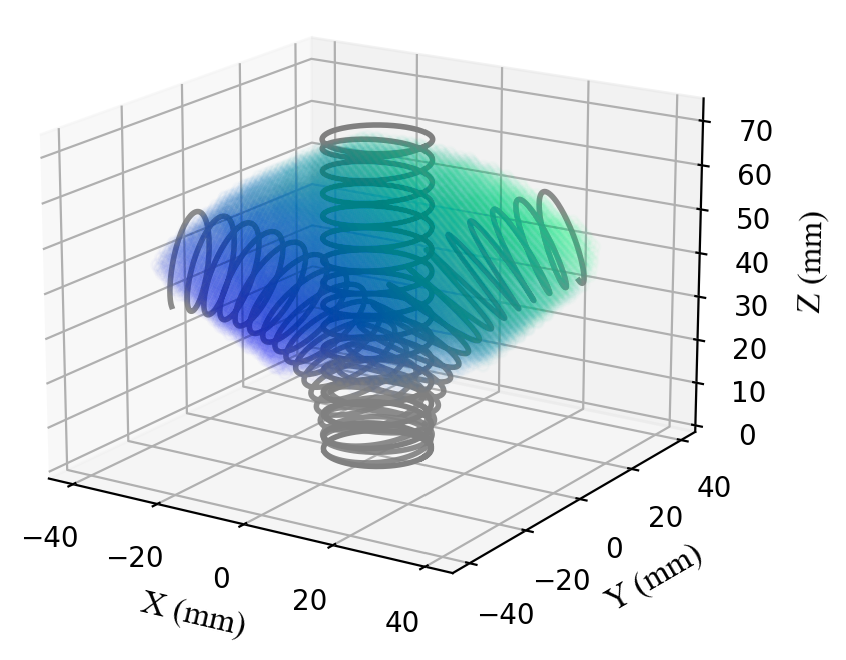}
  \caption{Workspace of the continuum robot. The coordinate system here is Frame $D$, and the origin is the center of the bottom of the spring.}
  \label{wrkspace}
\end{figure}

In the experiments, a spring with an outer diameter of 21 mm, an inner diameter of 18 mm, and a total length of 75 mm is used. Of this length, 3 mm is wrapped in the lower holder and 2 mm is secured in the upper holder, leaving a 70 mm effective bending length. The spring constant is 220 N/m, chosen to match the motor torque requirements. Four micro-filament braided lines, assumed inextensible, serve as the tendons. The tendons are controlled by four MG996R servo motors, with a rotation range of 0-120$^\circ$, each driving 3D-printed pulleys with a 7 cm diameter. Under the motor torque budget, the spring can be compressed to 20 mm (excluding the holder-embedded length). The simulated workspace, constrained by the spring length and torque limits, is shown in Fig. \ref{wrkspace}. For the artificial bristle, an LPS33HW sensor is used, and a 5.3 cm plastic filament works as the bristle. The motors and bristle sensor interface with an Arduino UNO board for communication and control, which connects to a host computer for command input and data logging.

\subsection{Validation of the Actuation Kinematic Model}

\begin{figure}[thpb]
  \centering
  \includegraphics[scale=0.48]{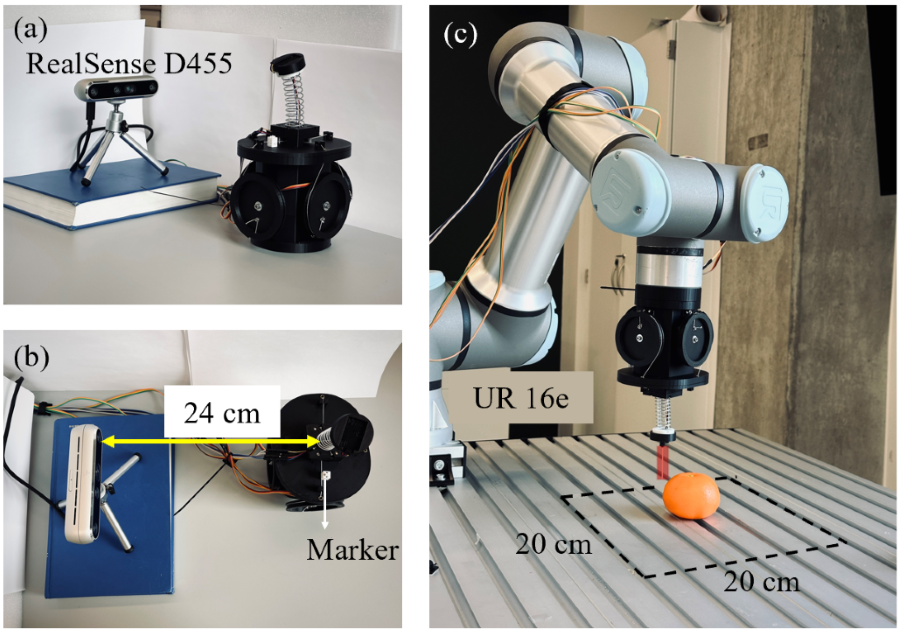}
  \caption{Experiment setups. (a) and (b) are the side view and the top view of the setup of the kinematics validation experiment. (c) is the setup for the object surface perception experiment.}
  \label{expSetup}
\end{figure}

This experiment verifies the kinematic model under both bending and compressing motion. The experiment setup is shown in Fig. \ref{expSetup} (a) and (b). A RealSense D455 camera (Intel) is placed 24 cm from the robot, and the camera lens is parallel to the robot bending plane. RGB images are recorded, and the robot positions is extracted in ImageJ. A 15-mm marker placed in the bending plane is used for scale calibration, and pixels are converted to spatial coordinates in ImageJ.

Given different Z-axis coordinates, the robot bends in either the X-Z plane or the Y-Z plane with bending angles $\theta$. Five measurement sets were conducted. The first evaluates pure compression ($\theta$=0). Nine length commands from 30 mm to 70 mm were given in 5 mm increments. The other four sets evaluate coupled bending and compression. For each set, seven Z coordinates were given, ranging from 40 mm to 70 mm with a step of 5 mm. Also, the angle $\theta$ ranged from $10^\circ$ to $90^\circ$ with a step of $10^\circ$. Based on the Z-axis coordinate and the angle $\theta$, the required length of the continuum robot was computed. The model considers only the 70 mm effective bending segment. Thus, 34 positions were tested for each direction (+X, -X, +Y, and -Y) in each set, with 136 locations in total for coupling motion tests.

\begin{figure}[thpb]
  \centering
  \includegraphics[width=0.47\textwidth]{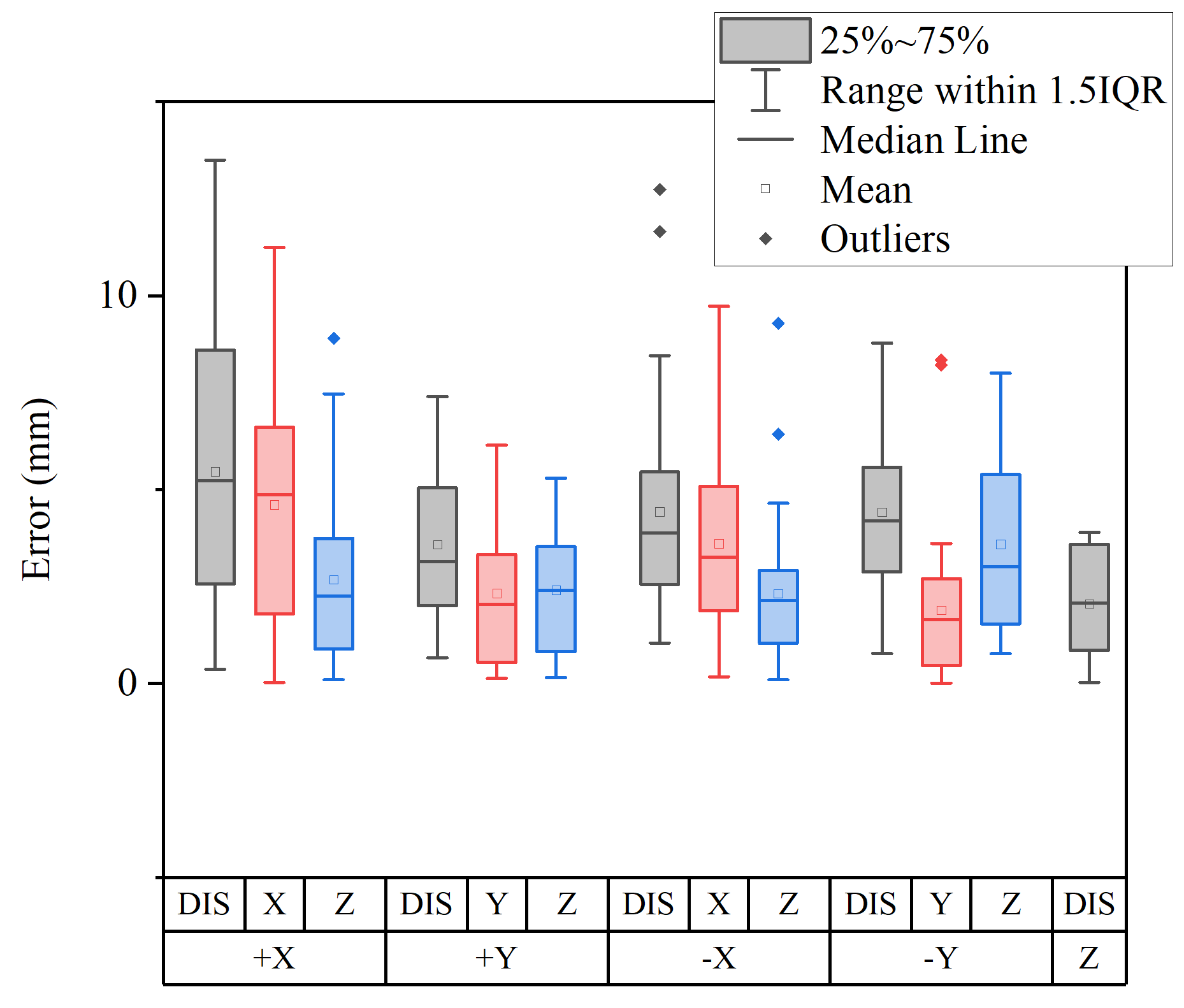}
  \caption{Box plot of position errors between the desired robot positions and the actual robot positions for the five sub-experiments. IQR is short for inter-quartile range, and DIS is short for displacement.}
  \label{boxPlot}
\end{figure}

Across 145 target positions, the mean position error is 4.32 mm with a standard deviation of 2.73 mm. For the pure compression test, the mean error is 2.06 mm with a standard deviation of 1.52 mm. Errors for the four coupled bending-compression tests are summarized in Table \ref{posError}, where SD denotes standard deviation and DIS denotes Euclidean distance. Axis-wise errors in $x/y$ and $z$ are also reported. Box plots for all five tests are shown in Fig. \ref{boxPlot}.

\begin{table}[th]
\caption{Errors on Coupling Motions of Robot Bending Toward Four Different Directions While Compressing}
\label{posError}
\begin{center}
\begin{tabular}{|c|c|c|c|c|c|c|}
\hline
~& \multicolumn{3}{|c|}{+X} & \multicolumn{3}{|c|}{-X} \\
\cline{2-7}
~ & DIS & X & Z & DIS & X & Z  \\
\hline
Mean (mm)& 5.47 & 4.61 & 2.68 & 4.43 & 3.60 & 2.31 \\
\hline
SD (mm) & 3.48 & 3.02 & 2.12 & 2.73 & 2.31 & 1.84  \\
\hline
\hline
~ & \multicolumn{3}{|c|}{+Y} & \multicolumn{3}{|c|}{-Y} \\
\cline{2-7}
~ & DIS & Y & Z & DIS & Y & Z   \\
\hline
Mean (mm)& 3.58 & 2.32 & 2.41 & 4.42 & 1.90 & 3.59 \\
\hline
SD (mm) & 1.86 & 1.69 & 1.49 & 2.23 & 1.92 & 2.08 \\
\hline
\end{tabular}
\end{center}
\end{table}

\begin{figure}[thpb]
  \centering
  \includegraphics[width=0.485\textwidth]{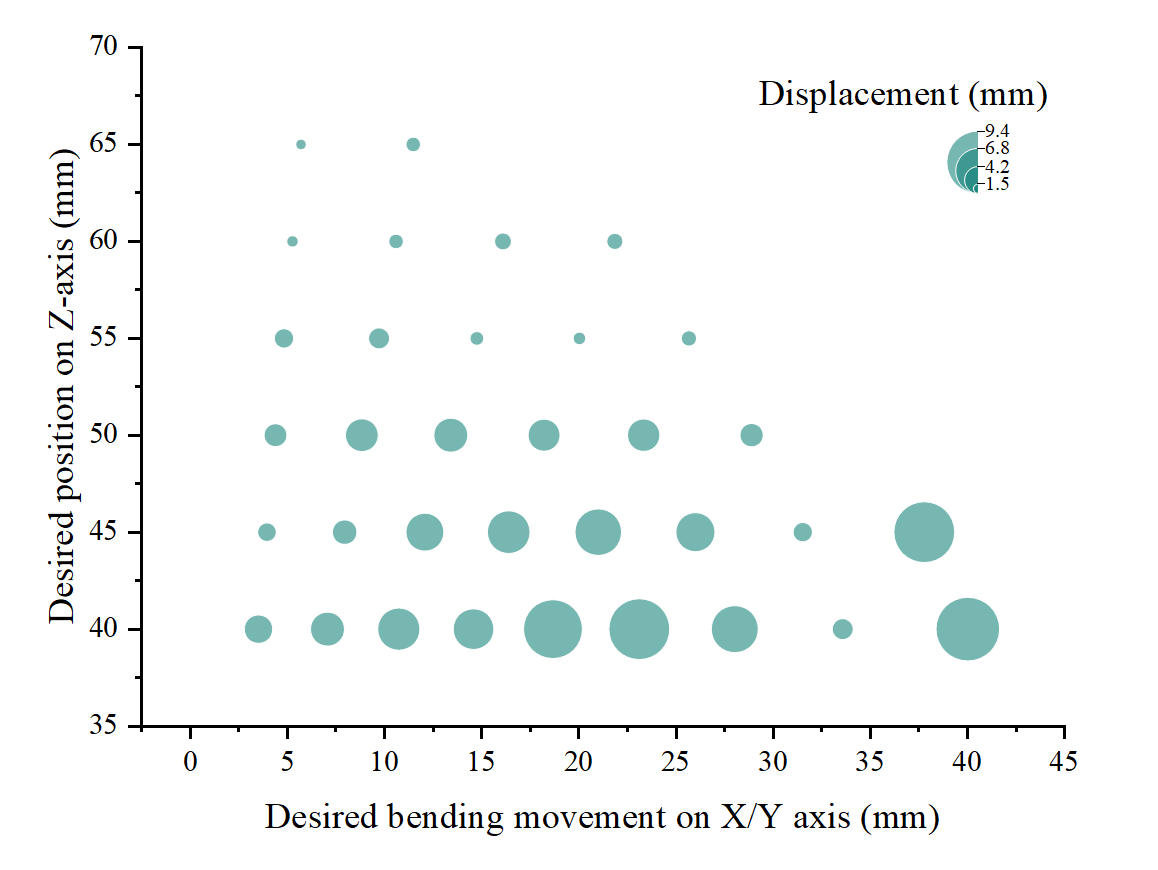}
  \caption{Bubble plot of the average errors at the 34 positions over the four sub-experiments of the coupling motions. The size of each bubble indicates the average error at the corresponding position.}
  \label{bubble}
\end{figure}


The robot achieves lower error in pure compression than in coupled motion, consistent with the simpler deformation. The errors in pure compressing may result from tendon displacement during winding. During the experiments, the tendons showed slight sliding along the winding paths, which may have caused displacement errors. Further improvements could be accomplished by updating the mechanical design and limiting tendon motion to eliminate such displacements. 

For coupled bending and compression, the mean position error is 4.32 mm. This error is likely due to the simplified kinematic model. For computational efficiency, tendon routing inside the body is approximated as either a constant-curvature arc or a straight segment. In practice, the tendon follows a shortest path between the two attachment points within the available internal space, and the mismatch can lead to control error. To further examine the coupled-motion results, Fig. \ref{bubble} shows a bubble plot of the mean errors at the 34 target positions. The horizontal axis represents the desired bending movement along the X or Y axis, and the vertical axis indicates the desired position along the Z-axis. Bubble size denotes the average errors of the four-directions at the the same commanded Z location and the same commanded movement distance along either the X or Y axis. Errors are larger in the middle region, which may indicate that the backbone shape in this range deviates from a circular arc, violating the constant-curvature assumption. The relatively large errors on the right side occurred when the spring shape was nearly quadrant. In this configuration, the tendons, assumed to be straight links due to their attached points on the spring's outer semicircle, were observed to slightly contact the inner spring surface, introducing additional deviation. 

\subsection{Object Surface Perception}

The goal of this experiment is to evaluate shape-based object surface perception from contact-driven point clouds obtained from the bristle feedback and the motions of both the continuum robot and a robot arm. As shown in Fig. \ref{expSetup} (c), a UR 16e industrial robot from Universal Robots is used only as a planar positioning stage that executes a repeatable X-Y plane scan (20 cm $\times$ 20 cm, zig-zag, 1 cm step). The continuum end-effector performs compliant vertical probing and contact detection via the artificial bristle, where contact is detected by an absolute pressure change of 15 hPa,  selected from pilot data by separating contact signals from non-contact pressure fluctuations. Originally, the continuum robot is fully extended. When the artificial bristle detects a contact, the continuum robot retracts to the minimum length to detach from the object. Then it extends the body down for the next contact, in which the extending length is utilized to measure the distance between the object and the robot. This strategy avoids lateral dragging-induced slip of the bristle filament and yields repeatable contact events for threshold-based detection, at the cost of reduced scanning speed. Once the distance is measured, the continuum robot retracts again before the UR robot moves to the next position. After the movement of the UR robot, the continuum robot extends again to explore the new position. These contraction and extension motions of the continuum robot repeat until no contact with the object is detected. In this way, the upper surface of the object can be roughly reconstructed with the collected point cloud information.

\begin{figure*}[thpb]
  \centering
  \includegraphics[width=\textwidth]{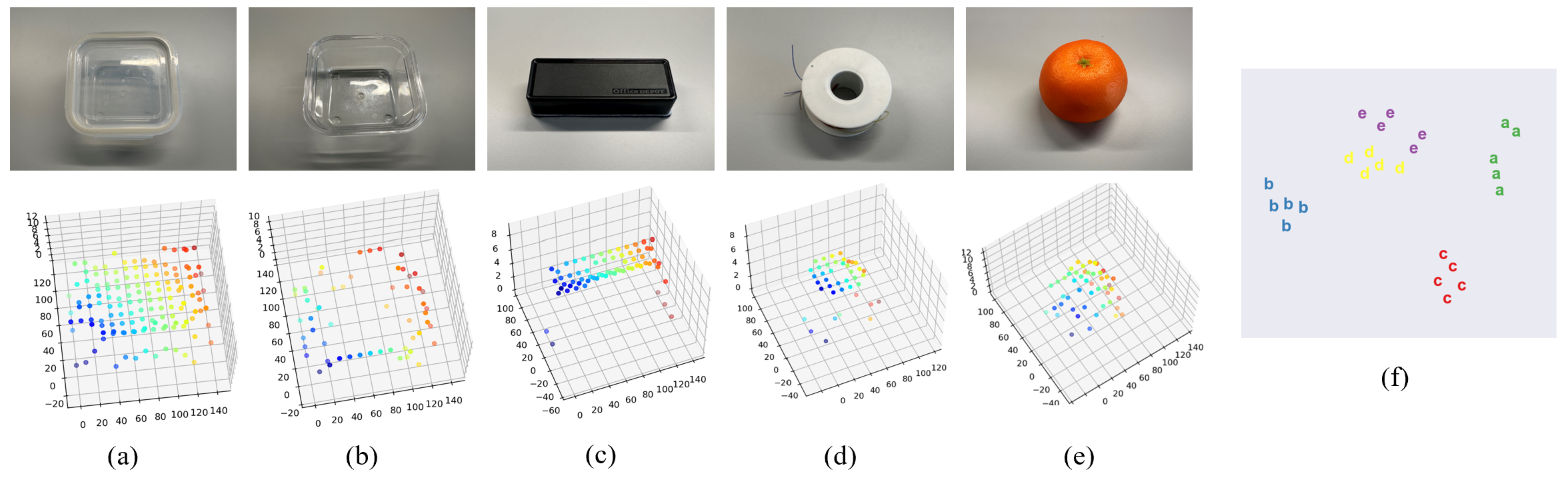}
  \caption{Five objects: (a) a food container with a lid, (b) a food container without a lid, (c) a board eraser, (d) a wire coil, and (e) an orange, are used in the object surface perception experiment. The figures on the upper row are the object images, and the figures on the lower row are the reconstructed upper surface point clouds of the objects. The color of the point indicates the distance to the origin, with red being further and blue being closer. (f) The visualization of point cloud data clustering using t-SNE.}
  \label{pointCloud}
\end{figure*}

As shown in the upper row of Fig. \ref{pointCloud} (a-e), five objects are selected in this experiment, with different sizes and shapes. In each trial, an object was placed within the 20 cm $\times$ 20 cm area at a random position. Given the UR robot motion information and the contact data from the continuum robot, the contact point clouds were recorded. After removing the non-contacted space and centralizing the point cloud data, the reconstructed upper-surface images of the five objects are drawn in the lower row of Fig. \ref{pointCloud} (a-e). The fluctuations of the contact detection points at the edges of the objects were mainly due to the slippery of the bristle filament along object surfaces. Nevertheless, the reconstructed height maps capture distinct shape cues for the tested objects. To assess trial-level separability for this set of tested objects, five trials of data were collected for each object by using the above pipeline. The data were aligned by the contact point with the least X and Y values and down-sampled to 20$\times$15 data points. Then the data were flattened and clustered using t-distributed stochastic neighbor embedding (t-SNE), as shown in Fig. \ref{pointCloud} (f). The embedding shows that trials from the same object form compact groups and different objects are separated in this dataset, suggesting that the proposed contact-driven surface perception pipeline captures object-dependent shape features under the tested conditions.

\subsection{Confined Space Exploration}

\begin{figure}[thpb]
  \centering
  \includegraphics[width=0.485\textwidth]{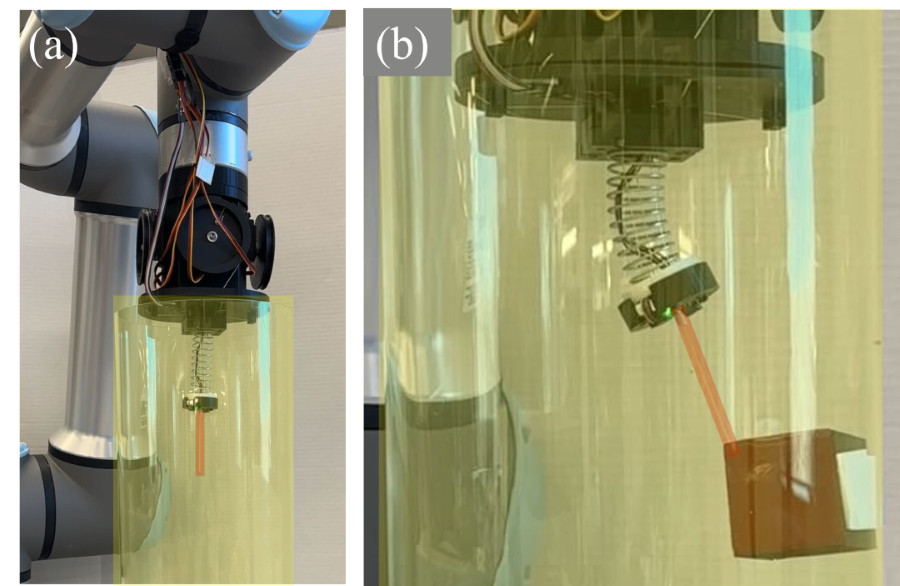}
  \caption{(a) Experimental setup for the confined space exploration task in a tube. (b) The bristle made contact with the obstacle cube.}
  \label{confinedSpace}
\end{figure}

A confined space exploration experiment was conducted to evaluate the ability of the spring-based continuum robot to detect obstacles within a constrained environment. The continuum end-effector was mounted on a UR16e robotic arm and used to scan the interior of a tube with an inner radius of 17.4 cm (Fig.~\ref{confinedSpace}), which was selected to accommodate the forearm-sized holder while constraining the holder-sized scanning envelope. The UR16e provided axial (z) positioning, while the continuum end-effector performed radial scanning through bending and contraction motions. In this setting, a detected contact indicates an occlusion that obstructs further insertion of the robot, and then the system retracts to avoid collision and the downward advance is terminated.

For the exploration task, the UR16e robotic arm moved downward in increments of 2 cm, performing five steps in total. After each step, the continuum robot first compressed to a length of 45 mm before extending to predefined target points $(15c_\alpha, 15s_\alpha,60)$ mm in Frame $D$ (Fig. \ref{knm}), where $\alpha \in [0, 2\pi)$ with increments of $\frac{\pi}{4}$. This allowed the bristle of the continuum robot to systematically scan the inner surface of the tube. The scanning process followed a cyclic sequence: compression, extension to a specific $\alpha$ angle, re-compression, and subsequent extension to the next target position. The motion of the continuum robot was smoothed by interpolating tendon length changes to ensure continuous movement. During scanning, contact was detected based on a pressure variation threshold of 15 hPa. If contact was registered while the continuum robot was extending, it would immediately stop extending, retract to the compressed state, and continue scanning in the next direction until the scanning process was complete. If any contact was detected during the scan, the UR16e robotic arm would return to its initial position after the scan rather than continuing its downward movement.

To introduce obstacles in the environment, a cubic object with an edge length of 4 cm was attached inside the tube at four different vertical distances from the compressed bristle tip: 3.5 cm, 5.5 cm, 7.5 cm, and 9.5 cm. In each trial, the robotic arm stopped descending at 4 cm, 6 cm, 8 cm, and 10 cm, respectively, upon successful detection of the obstacle. For comparison, a control experiment was conducted without an obstacle inside the tube. In this condition, the robotic arm descended a total of 10 cm, and no contact was detected during any scan. This experiment demonstrated the feasibility of using the spring-based continuum robot for confined space exploration and obstacle detection.

\section{CONCLUSION}
Inspired by the compressive bending movement of caterpillars, this paper introduces the development of a spring-based tendon-driven continuum robot with enhanced flexibility and dexterity. The tight actuation module design enables easy integration with commercial robots as an end effector, enhancing versatility and adaptability in various applications. With a constant curvature kinematic model, this continuum robot is controlled by the tendon lengths and can reach the desired position through both bending and compressing motions with an average error of 4.32 mm. Furthermore, when combined with an artificial bristle, this compliant and non-invasive end effector demonstrates remarkable capabilities in performing inspection tasks, such as object surface perception in a confined space. This system provides a low-cost option for upgrading existing commercial robots with simple mechanical components for exploration in constrained environments. The proposed design is not intended for load-bearing manipulation. Payload capacity and maximum tip force are limited and not characterized in this work. Future work will focus on improving positioning accuracy through calibration and compensation and closed-loop sensing to account for backbone compressibility and unmodeled tendon effects.

\addtolength{\textheight}{-8cm}   





\bibliographystyle{IEEEtran}
\bibliography{IEEEexample}

\end{document}